\documentclass[letterpaper, 10 pt, conference]{ieeeconf}

\IEEEoverridecommandlockouts               
\renewcommand{\baselinestretch}{0.98}

\usepackage[T1]{fontenc}
\usepackage{amssymb,amsmath,epsfig}
\usepackage[latin5]{inputenc}
\usepackage{graphicx,times,latexsym}
\usepackage{subfigure}
\usepackage{psfrag,cite}
\usepackage{setspace}
\usepackage{algorithmic, algorithm}
\usepackage{color}
\usepackage{url}

\usepackage{mathtools}
\mathtoolsset{showonlyrefs=true}

\allowdisplaybreaks
 
\title{Path Planning and Controlled Crash Landing of a Quadcopter \\ in case of a Rotor Failure}

\author{Mojtaba Hedayatpour, Mehran Mehrandezh, Farrokh Janabi-Sharifi 
\thanks{M. Hedayatpour is with DOT Technology Corp. Regina, Canada. M. Mehrandezh is with the Faculty of Engineering \& Applied Science, University of Regina, Canada. F. Janabi-Sharifi is with the Department of Mechanical \& Industrial Engineering, Ryerson University, Canada. 
\newline mojtaba@seedotrun.com, mehran.mehrandezh@uregina.ca, \newline fsharifi@ryerson.ca}}

\begin{document}
\maketitle
\begin{abstract}
This paper presents a framework for controlled emergency landing of a quadcopter, experiencing a rotor failure, away from sensitive areas. A complete mathematical model capturing the dynamics of the system is presented that takes the asymmetrical aerodynamic load on the propellers into account. An equilibrium state of the system is calculated around which a linear time-invariant control strategy is developed to stabilize the system. By utilizing the proposed model, a specific configuration for a quadcopter is introduced that leads to the minimum power consumption during a yaw-rate-resolved hovering after a rotor failure. Furthermore, given a 3D representation of the environment, an optimal flight trajectory towards a safe crash landing spot, while avoiding collision with obstacles, is developed using an RRT* approach. The cost function for determining the best landing spot consists of: (i) finding the safest landing spot with the largest clearance from the obstacles; and (ii) finding the most energy-efficient trajectory towards the landing spot. The performance of the proposed framework is tested via simulations. 
\end{abstract}

\section{Introduction}\label{sec:intro}
Multicopters have gained attention in recent years. Due to their simplicity and maneuverability, they have been used in a broad spectrum of applications such as agronomy~\cite{weed}, calibrating antenna of a telescope~\cite{antenna} and inspection of infrastructures~\cite{avian}. 

A special type of multicopters with four motors, known as quadcopters, has been extensively studied and there is a vast literature about their modeling, design, control and path planning. These vehicles normally have an even number of propellers half of which turn in the opposite direction of the remaining propellers. Modeling and full control of a quadcopter can be found in~\cite{bouabd}.

Fault tolerant control of multicopters in case of partial or complete failure of actuators is an area of interest among researchers. Feedback linearization approach is used in~\cite{freddi} to stabilize a quadcopter after complete loss of one propeller. Stability and control of quadcopters experiencing one, two or three rotor failures are presented in~\cite{muller}, however all propellers have parallel axes of rotation and the effects of rotation of center of mass of the propellers on their performance are not investigated. To increase reliability by redundancy, quadcopters with tilting rotors, hexacopters and octacopters are introduced which are capable of maintaining stable flight despite losing one to four actuators~\cite{ryll,scaramutza,ukass}, however they are not optimal in terms of power consumption or stability. Emergency landing for a quadcopter with one rotor failure in an environment without obstacles can be found in~\cite{serra} where the landing location is known and path planning is not discussed. 

An emerging area of research in multi-rotor UAVs falls into finding a landing spot and planning a safe trajectory towards it in case of rotor failure. While there is a huge body of literature on trajectory planning of quadcopter~\cite{mellinger,hehn,lavalle,sertac}, the work done on extending this to a situation where there is a rotor failure is scarce. 
Despite the unprecedented progress in the development of UAVs (especially multicopters) in recent years, two major issues, namely safety and endurance, still remain as main challenges. These vehicles are prone to having different types of failure in the system such as partial or complete loss of motors or propellers, collision with obstacles or other vehicles and power outage. Since they are becoming an inevitable part of our everyday life, safety becomes one of the key factors in designing such vehicles.

This paper presents a framework for emergency landing of a quadcopter in case of a rotor failure. Because of the fast rotations of the vehicle's body (due to unbalanced moments in the system), it is essential to consider all aerodynamic effects. In particular, two important parameters affecting propeller's performance, namely the resultant angular velocity of the propeller and the freestream velocity, are investigated for the first time. Based on blade element theory~\cite{anderson} a complete mathematical model for the propeller is incorporated in equations of motion which results in finding a specific configuration of quadcopters namely, adding a tilting angle to the rotors thrust vector, which leads to the minimum power consumption in hovering. Hover solution for different configurations is calculated and a comparison in terms of power consumption amongst them is presented. For the configuration with minimum power consumption, cascaded control strategy is used to control attitude and position of the vehicle and nonlinear simulations validating the results are presented. 

For completing the landing, first an algorithm is proposed to find the best landing spot in a given map of the environment where obstacles are represented by cuboids. Two parameters are used to define a cost function in order to find the optimal landing spot in the given map: (i) finding the safest landing spot with the largest clearance from the obstacles; and (ii) finding the most energy-efficient trajectory towards the landing spot. In order to properly define the clearance from obstacles, Generalized Voronoi Diagram (GVD) is used. For all points on the GVD, the one with minimum cost is selected as the landing spot. A finite horizon is selected in generating the GVD. The boundary of this horizon is estimated based on the total cost-to-go based on the power requirement. Furthermore, due to the size and dimensionality of the search space, an RRT*-type randomized motion planning strategy is adopted that can generate optimal trajectories on the fly in real time~\cite{sertac}. Using nonlinear simulations and the designed controller, the results of following the path and performing emergency landing are evaluated. 


The paper is organized as follows. In Section~\ref{sec:model}, mathematical modeling of a quadcopter with a complete aerodynamic model of a propeller in presence of freestream velocity is presented. Equilibrium state after failure, fault tolerant control design, effects of tilting the rotors on power consumption and introduction of a specific configuration with minimum-power hover solution are presented in Section~\ref{sec:control}. In section~\ref{sec:path}, an algorithm to find the best landing spot and path planning using sampling-based planning algorithms are presented. Finally, nonlinear simulation results are presented in Section~\ref{sec:sim} and the paper concludes in Section~\ref{sec:conc}. 

\subsubsection*{Notation}
Matrices are represented by straight boldface letter and all vectors are represented by italicized boldface letters. Rotation matrix between frame i and frame j is represented by $^j\textbf{R}_i$ . In addition, the term $^I\omega_B$  denotes that $\omega$ belongs to $B$ and is expressed in frame $I$. Angular velocity vector of the vehicle is represented by $\pmb{\omega}^B=(p,q,r)^T$ where $p$, $q$ and $r$ are roll, pitch and yaw rates respectively. Finally, $\|\pmb{\omega}\|$ represents the 2-Norm of the vector $\pmb{\omega}$ and $|s|$ represents the absolute value of scalar $s$.

\section{Modeling}\label{sec:model}
In this section a complete dynamic model of the quadcopter is given, considering the aerodynamic model of the propellers' thrust against freestream velocity. 
\begin{figure}[t]
	\centering
 \includegraphics[scale=0.63]{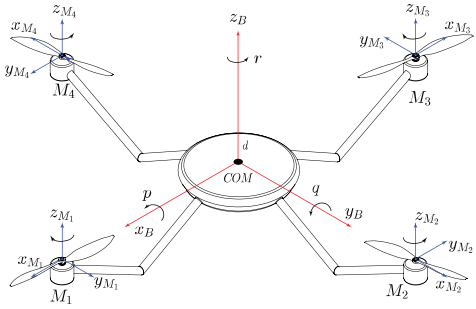}
	\caption{Quadcopter in $+$ configuration. }\label{fig:quad} 
\end{figure}

\subsection{Equations of Motion}
Figure~\ref{fig:quad} shows schematic of a quadcopter. Six reference frames are defined, one of which is assumed to be fixed and attached to the earth, also known as inertial frame $I$, one is attached to the center of mass of the vehicle that rotates with it and is represented by $B$ and four other reference frames attached to the center of mass of the $i^{th}$ motor $M_i$, however they do not turn with the rotors. 

A propeller is attached to each motor equipped with a propeller which generates a thrust force $f_i$ in the direction of z-axis of the motor frame. Propellers 1 and 3 have negative angular velocity and propellers 2 and 4 have positive angular velocity expressed in the body frame as $\pmb{\omega}^{p_i}=(0,0,\omega^{p_i} )^T$. The moment of inertia of the propellers is approximated by the moment of inertia of a disk and is represented by a diagonal matrix as $\textbf{I}^p=\textrm{diag}(I_{xx}^p,I_{yy}^p,I_{zz}^p )$ . The angular velocity of the body frame with respect to the inertial frame is represented by $\pmb{\omega}^B=(p,q,r)^T$. The geometry of the vehicle is assumed to be symmetric so its moment of inertia matrix can be represented by a diagonal matrix as $\textbf{I}^B=\textrm{diag}(I_{xx},I_{yy},I_{zz})$. The equations of motion are: 
\begin{align}
    \label{eq:rot_motion}
    \textbf{I}^B \dot{\pmb{\omega}}^B +
    \sum_{i=1}^4\textbf{I}^p\dot{\pmb{\omega}}^{p_i} + \\
    \textrm{sk}(\pmb{\omega}^B) \Big( \textbf{I}^B \pmb{\omega}^B + \sum_{i=1}^4\textbf{I}^p(\pmb{\omega}^{p_i}+\pmb{\omega}^B) \Big) 
    = \\
    \pmb{\tau}_{\textrm{lift}} + \pmb{\tau}_d + \pmb{\tau}_{\textrm{reaction}} + \pmb{\tau}_p,
\end{align}
\begin{align}
    \label{eq:trans_motion}
    m\Ddot{\pmb{d}} = ^I\textbf{R}_B\pmb{f} + m \pmb{g} + \pmb{f}_d.
\end{align}

In the left hand side of~\eqref{eq:rot_motion}, the first and second terms are moments due to angular accelerations of body and propellers. The third term represents cross-coupling of angular momentum because of the rotation of the body and propellers and $\textrm{sk}(\pmb{\omega}^B)$ represents the skew-symmetric matrix of the angular velocity of the body. In the right hand side of~\eqref{eq:rot_motion}, the first term is the moment due to propeller's thrust force, the second term is the moment due to drag force of the body, the third term is the reaction moment of the propeller and the last term is the moment due to asymmetrical lift distribution over the advancing and retreating blades of the propellers~\cite{csme_paper}. The moment due to aerodynamic drag, $\pmb{\tau}_d$, is assumed to be proportional to angular velocity of the vehicle $\pmb{\omega}^B$ with a proportionality constant $\textbf{K}_d=\textrm{diag}(k_{dx},k_{dy},k_{dz})$. The reaction moment of the propeller is assumed to be proportional to the thrust force of the propeller $\pmb{f}_p$, with a constant $k_\tau$~\cite{ktau}. The derivation of $\pmb{\tau}_p$ and $\pmb{f}_p$ are presented in our previous work~\cite{csme_paper}. 
In~\eqref{eq:trans_motion}, the position of the quadcopter's center of mass in the inertial frame is denoted by $\pmb{d}=(d_1,d_2,d_3)$. In the right hand side, $\pmb{f}$ is the sum of all the forces generated by propellers as expressed in the body frame, $\pmb{f_d}$ is the aerodynamic drag force due to translational motion of the fuselage and is assumed to be proportional to the linear velocity of the center of mass of the vehicle with a proportionality constant $\textbf{K}_D=\textrm{diag}(k_{Dx},k_{Dy},k_{Dz})$, $\pmb{g}$ is the gravitational acceleration and $^I\textbf{R}_B$  is the rotation matrix from body frame to inertial frame. 

\section{Control Design For a Quadcopter Experiencing One Rotor Failure}\label{sec:control}
In this section a new stable hovering definition for a quadcopter experiencing one rotor failure is presented first. Second, based on this new hovering definition, a control strategy is developed to control attitude and position of the vehicle after failure. Finally, it is shown how by titling the rotors one can minimize the power consumption and also improve the flight stability in case of one rotor failure. 

\subsection{Stable Hovering Definition In Case of Rotor Failure}\label{subsec:hoverdef}
Generally, in multi-rotor UAVs, hovering is defined as maintaining a position with zero angular and linear velocities. However, in case of one rotor failure in a quadcopter and in order to control the attitude and position of the vehicle, a new hovering definition is required. Hovering is defined as maintaining an altitude while rotating with constant angular velocity about a unit vector that is fixed with respect to the vehicle~\cite{muller}. 

Without loss of generality, suppose motor number 4 (see Fig. 1) is failed. Because of the unbalanced moments of the remaining functioning propellers, the vehicle starts rotating about a unit vector $\pmb{n}$ (as expressed in the body frame) with angular velocity $\pmb{\omega}^B$. The evolution of this unit vector in time can be written as follows: 
\begin{align}
    \label{eq:ndot_zero}
    \dot{\pmb{n}} = -\pmb{\omega}^B\times \pmb{n}.
\end{align}
According to the new hovering definition, we want this unit vector to be fixed with respect to the vehicle. If this unit vector is fixed, from~\eqref{eq:ndot_zero} we can say the angular velocity of the vehicle will remain parallel to this unit vector so the vehicle will be rotating about $\pmb{n}$. Now if $\pmb{\omega}^B$ remains constant, one can achieve stable hovering as all the states of the system will remain bounded. If $\pmb{n}$ is fixed,~\eqref{eq:ndot_zero} should be equal to zero. In other words, in hover, $\pmb{n}$ is a unit vector stationary in the inertial frame as expressed in the body frame which is parallel to $\pmb{\omega}^B$ vector. Setting~\eqref{eq:ndot_zero} to zero and knowing that $\pmb{n}$ is a unit vector, one can write the followings (note that an overbar indicates equilibrium values): 
\begin{align}
    \label{eq:sigma}
    \dot{\pmb{n}} = 0 \xrightarrow{} \|\Bar{\pmb{n}}\| = 
    \sigma \|\Bar{\pmb{\omega}}^B\| = 1 \xrightarrow{} \sigma =
    \frac{1}{\|\Bar{\pmb{\omega}}^B\|}.
\end{align}
Also, in hover, the projection of total thrust forces of all propellers onto $\Bar{\pmb{n}}$ should balance the weight of the vehicle which results in adding the following constraint to the system: 
\begin{align}
    \label{eq:weight_balance}
    \sum_{i=1}^4 \Bar{\pmb{f}}_{p_i} \cdot \Bar{\pmb{n}} = m \|\pmb{g}\|.
\end{align}

As the vehicle is turning with constant angular velocity $\Bar{\pmb{\omega}}^B=(\Bar{p},\Bar{q},\Bar{r})^T$, the center of mass of the $i^{th}$ propeller goes through a rotation about $\Bar{\pmb{n}}$  which generates a uniform freestream velocity $V_{\infty}=\Bar{r}l$ over the propeller (where $l$ is the distance of the center of mass of the propeller from the center of mass of the vehicle)~\cite{csme_paper}. Considering this freestream velocity, using the proposed propeller model in~\cite{csme_paper} and the resultant angular velocity of the propellers, total thrust force and moment of the $i^{th}$ propeller can be written as follows: 
\begin{align}
    \label{eq:prop_f}
    f_{p_i} = \rho_a c C_L \Big(
    \frac{R_b^3\omega^{{p_i}^2}}{3} + 
    \frac{R_b^3\Bar{r}^2}{3} + 
    \frac{R_b\Bar{r}^2l^2}{2} + 
    \frac{2R_b^3\Bar{r}\omega^{p_i}}{3}
    \Big)\:\:\:
\end{align}
\begin{align}
    \label{eq:prop_tau}
    \tau_{p_i} = \rho_a c C_L \Big(
    \frac{R_b^3\Bar{r}l\omega^{p_i} + R_b^3\Bar{r}^2l}{3}
    \Big),
\end{align}
where $\rho_a$ is the air density, $c$ is propeller's blade chord, $C_L$ is propeller's lift coefficient and $R_b$ is the propeller's blade radius. Using equations~\eqref{eq:rot_motion}-~\eqref{eq:ndot_zero}, by setting angular accelerations to zero  and considering the proposed propeller model, a system of eight algebraic equations for 11 unknowns are obtained. Three more equations are required to solve the system. The unknowns are: $\Bar{p}$, $\Bar{q}$, $\Bar{r}$ , $\Bar{n}_x$, $\Bar{n}_y$, $\Bar{n}_z$, $\sigma$, $\Bar{\omega}^{p_1}$, $\Bar{\omega}^{p_2}$, $\Bar{\omega}^{p_3}$, $\Bar{\omega}^{p_4}$. Assuming that motor number 4 is failed ($\Bar{\omega}^{p_4}=0$) and by adding the following constraints, we will end up with a system of 11 algebraic equations with 11 unknowns. 
\begin{align}
    \label{eq:add_constraints}
    \Bar{\omega}^{p_1}=\Bar{\omega}^{p_3} \: \: , \:\:
    \rho = \Big( \frac{\Bar{\omega}^{p_2}}{\Bar{\omega}^{p_1}} \Big)^2,
\end{align}
where $\rho$ is a tuning factor and a non-negative scalar. Now there are 11 algebraic equations to be solved for 11 unknowns to obtain equilibrium values or hover solution. 
For simplicity, assuming  $\textbf{I}^p \ll \textbf{I}^B$, one can neglect the second term in~\eqref{eq:rot_motion}. Also, since yaw is the dominant rotational motion, $\pmb{\tau}_d$ is assumed to oppose yaw motion only and is assumed to be proportional to yaw rate with proportionality constant $\beta$ as $\pmb{\tau}_d = (0,0,-\beta r)^T$. The reaction moment of the propeller is also assumed to be proportional to its thrust force and can be expressed in the body frame as $\pmb{\tau}_{\textrm{reaction}, p_i} = - \textrm{sign}(\omega^{p_i})k_{\tau}\pmb{f}_{p_i}$. Therefore, using the reaction torque and the angular velocity of the propeller, the power consumption of the motors in hover can be calculated as follows:
\begin{align}
    \label{eq:motor_power}
    \Bar{P}_{p_i} = \Bar{\tau}_{\textrm{reaction}, p_i} \Bar{\omega}^{p_i} = 
    - \textrm{sign}(\omega^{p_i})k_{\tau} \|\pmb{f}_{p_i}\| \omega^{p_i}.
\end{align}

\subsection{Control Design}\label{subsec:control}
In this section, control design is presented. A cascaded control strategy  is used to control attitude and position of the vehicle. Using nonlinear equations of rotational motion in~\eqref{eq:rot_motion}, a linear time-invariant system is introduced to control the attitude of the vehicle or in other words control the direction of the unit vector $\pmb{n}$. In addition, it is shown that by controlling two rotational degrees of freedom that are related to the translational motion ($n_x$ and $n_y$), along with the sum of all thrust forces, the position of the vehicle can be controlled. 

In controlling the attitude, the strategy is to give up control of the full attitude after failure. Instead, only those rotational degrees of freedom  related to translational motion of the vehicle will be controlled which is often called reduced attitude control~\cite{cdsr_paper}. After failure, reduced attitude states are represented by a variable $\pmb{\zeta}=(p,q,n_x,n_y)$ which includes pitch and roll rates of the vehicle and $x$ and $y$ components of the unit vector $\pmb{n}$. By linearizing~\eqref{eq:rot_motion} and~\eqref{eq:ndot_zero} about the equilibrium state $\Bar{\pmb{\zeta}} = (\Bar{p},\Bar{q},\Bar{n}_x,\Bar{n}_y)$, the deviations of $\pmb{\zeta}$ from $\Bar{\pmb{\zeta}}$ as represented by $\Tilde{\pmb{\zeta}}$ can be described by the following linear time-invariant system: 
\begin{align}
    \label{eq:reduced_atti}
    \dot{\Tilde{\pmb{\zeta}}} = A\Tilde{\pmb{\zeta}} + B\pmb{u} 
    \:\: , \:\: 
    A = \frac{\partial\dot{\pmb{\zeta}}}{\partial\pmb{\zeta}}\Big|_{\pmb{\zeta}=\Bar{\pmb{\zeta}}}
    \:\: , \:
    B = \frac{l}{I_{xx}}
    \begin{bmatrix}
    0 & 1 \\
    1 & 0 \\
    0 & 0 \\
    0 & 0 
    \end{bmatrix}, \:\:\:\:\:\:\:\: \\
    u_1 = \|(\pmb{f}_{p_3}-\Bar{\pmb{f}}_{p_3}) - (\pmb{f}_{p_1}-\Bar{\pmb{f}}_{p_1}) \|
    \:,
    u_2 = \|(\pmb{f}_{p_2}-\Bar{\pmb{f}}_{p_2})\|
\end{align}
Also in equilibrium, to balance the weight of the vehicle can be determined by the following constraint: 
\begin{align}
    \label{eq:balance_constraint_hover}
    \pmb{f}_{p_1}+\pmb{f}_{p_2}+\pmb{f}_{p_3} = \Bar{\pmb{f}}_{p_1}+\Bar{\pmb{f}}_{p_2}+\Bar{\pmb{f}}_{p_3}.
\end{align}
By designing a linear controller for~\eqref{eq:reduced_atti}, the inner loop of the cascaded controller is complete. 

In order to control the position of the vehicle, an outer control loop is designed such that it generates reference signal for the inner control loop. This can be done first by finding the desired acceleration of the vehicle to get to the desired position and then transforming it to the desired direction of the unit vector $\pmb{n}$. The desired acceleration can be found by defining a new state variable $\Tilde{\pmb{d}}$ as expressed in the inertial frame, as the deviations of the position of the vehicle d from its desired position $\pmb{d}_{des}$, behaving like a second order system with damping ratio $\xi$ and natural frequency $\omega_n$ as follows: 
\begin{align}
    \label{eq:acc_des}
    \Ddot{\pmb{d}}_{des} = -2\xi\omega_n \dot{\Tilde{\pmb{d}}} - \omega_n^2\dot{\Tilde{\pmb{d}}}.
\end{align}
The total acceleration is then defined as $(\Ddot{\pmb{d}}_{des}-\pmb{g})$. In hover, we want $\Ddot{\pmb{d}}_{des}=0$ so that the desired direction of the unit vector $\pmb{n}$ will be in the opposite direction of $\pmb{g}$. According to the Newton's second law, one can write the following equation to find the desired direction of $\pmb{n}$: 
\begin{align}
    \label{eq:n_des}
    \Big(\sum_{i=1}^4 {^B\pmb{f}_{p_i}} \cdot \Bar{\pmb{n}}\Big) \pmb{n}_{des} =
    m {^I\textbf{R}_B}(\Ddot{\pmb{d}}_{des}-\pmb{g}).
\end{align}
In summary, the outer control loop controls the position of the vehicle and generates reference signal for the inner control loop which controls the reduced attitude of the vehicle. 

\subsection{Effects of Tilting The Rotors On Power Consumption}\label{subsec:tilting}
In previous sections, hover solution and control design were presented. In this section, effects of tilting the rotors on power consumption of the motors after failure are presented. 

After failure, it is shown that at equilibrium, the vehicle will have constant angular velocity with yaw being the dominant rotational motion. In hover, according to~\eqref{eq:prop_f} and~\eqref{eq:prop_tau}, $\Bar{r}$ can have a significant effect on thrust force and the moment generated by the propellers depending on its magnitude and direction which consequently can affect power consumption of the motors. In particular, a specific configuration can be introduced that generates $\Bar{r}$ such that it is in favor of thrust force and the moment of the propeller and thus yielding the minimum-power hover solution. 

In addition, in hover, $\Bar{r}$ can affect the resultant angular velocity of the propeller and also can change the relative air flow velocity over the blade. In a quadcopter, because half of the rotors are turning in the opposite direction of the remaining half of the rotors, therefore after failure, $\Bar{r}$ will have positive effect on some rotors and negative effect on some other rotors. If the direction of $\Bar{r}$ is the same (opposite) as that of the propeller's angular velocity, then the propeller should turn slower (faster) in order to generate the same amount of thrust force when $\Bar{r}=0$, therefore according to~\eqref{eq:motor_power}, since $\pmb{f}_{p_i}$ experiences very small changes (only for small angles), the power consumption of the motor will be decreased (increased). The goal of this section is to find the best configuration of the rotors to get the most benefits out of $\Bar{r}$ after failure, such that the power consumption of the motors is minimized. 

In regular quadcopters, yaw motion is usually carried out using the reaction moment of the propellers. This moment is fairly small compared to the moment generated by the propeller's thrust force about the center of mass of the vehicle [6], therefore it may not be an efficient way to yaw. Instead, one can yaw by tilting the rotors by angle $\alpha$ about the x-axis of their corresponding motor frame and using a small component of the propeller's thrust force to generate relatively larger yaw moments~\cite{muller}. Note that the tilting angle should be small enough so that~\eqref{eq:prop_f} and~\eqref{eq:prop_tau} can hold true and the component of the thrust force that balances the weight of the vehicle experiences fairly small changes. 

\begin{figure}[b]
	\centering
    \includegraphics[scale=0.44]{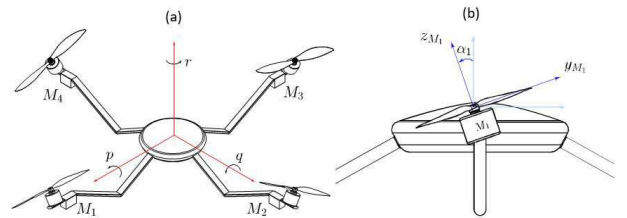}
	\caption{(a) a new configuration with tilted rotors. (b) the positive direction of the tilting angle $\alpha$.} 
	\label{fig:tilting} 
\end{figure}

A new configuration is proposed by tilting the rotors about the x-axis of the motors frame (shown in blue in Fig. 1) as shown in Fig.~\ref{fig:tilting} (a) where the positive direction of the tilting angle $\alpha_i$ is shown in Fig.~\ref{fig:tilting} (b). Because rotors 1 and 3 are assumed to be turning in the negative direction of z-axis of the body frame, by tilting these motors by any positive angle, the vehicle tends to generate a yaw motion that is in favor of reducing their power consumption. Whereas for rotors 2 and 4 which are turning in the positive direction of the z-axis of the body frame, the tilting angle should be negative. Note that, for simplicity, it is assumed $\alpha_1=\alpha_3$ and $\alpha_2=\alpha_4$.
This new configuration not only helps to reduce the power consumption after failure, but also helps to increase stability of the vehicle in yaw motion in absence of failures~\cite{iros_paper}. It also adds a new tuning parameter $\alpha_i$ to the hover solution. 

Assuming motor number 4 is failed, using~\eqref{eq:motor_power} to find the minimum-power hover solution a simple line search is performed over tuning parameters $\rho$ and $\alpha$. Results show that the minimum-power solution can be found when $\rho=0$ and $\alpha_{1,3}=0.4$  rad, meaning that  after failure, motor number 2 should be turned off and motors number 1 and 3 should be tilted by 0.4 rad. 

Next, to investigate the optimality of the hover solution, we calculate power consumption of the motors in hover for various scenarios. For this purpose, an example quadcopter with the following parameters is considered. 

Consider a vehicle with $m=0.5$ kg, $I_{xx}=I_{yy}=3.2\times10^{-3}$ kg.m$^2$, $I_{zz}=5.5\times10^{-3}$  kg.m$^2$, $l=0.17$ m, $I_{zz}^p=1.5\times10^{-5}$  kg.m$^2$, $\beta=2.75\times10^{-3}$ and $k_\tau=1.69\times10^{-2}$  m. The propellers have two blades with $c=0.03$ m, $C_L=1.022$, $R_b=0.08$ m and air density is assumed to be constant $\rho_a=1.225$ kg/m$^3$. For this vehicle, minimum-power hover solution can be found by setting $\rho=0$ and $\alpha_{1,3}=0.4$ as follows: 
\begin{align}
    \label{eq:min_power_hover_sol}
    \Bar{\pmb{n}} = (0,0,1)^T, \:\; \Bar{\omega}^{p_1}=\Bar{\omega}^{p_1} = -499.2\: \textrm{rad/s}, \\
    \Bar{\omega}^{p_2}=\Bar{\omega}^{p_4}=0, \: \Bar{\pmb{\omega}}^B = (0,0,-95.47)^T\: \textrm{rad/s},\\
    \Bar{f}_{p_1}=\Bar{f}_{p_3}=2.66 \: \textrm{N}, \: \Bar{f}_{p_2}=\Bar{f}_{p_4}=0, \\
    \Bar{\pmb{\tau}}_{p_1}=\Bar{\pmb{\tau}}_{p_3}=(0,-0.04,0)^T \: \textrm{N.m}, \: \Bar{\pmb{\tau}}_{p_2}=\Bar{\pmb{\tau}}_{p_4}=0, \\
    \Bar{P}_{\textrm{hover}}=\sum_{i=1}^4\Bar{P}_{p_i} = 44.9 \: \textrm{W}.
\end{align}

Note that if we set the tilting angle to zero and minimize the total power consumption in hover with respect to $\rho$, the minimum power would be equal to $\Bar{P}_{\textrm{hover}}=46.25$ W. Furthermore, if we set both tilting angle and $\rho$ to zero, the minimum power becomes  $\Bar{P}_{\textrm{hover}}=54$ W. 

\section{Path Planning For Crash Landing}\label{sec:path}
As the goal of the paper is to introduce a framework for crash landing, in this section, we continue by presenting path planning and crash landing a quadcopter after a rotor failure. 
Recently, sampling-based planning algorithms such as Rapidly-exploring Random Trees (RRT) have proved to be practical and effective in high-dimensional state spaces and have attracted considerable attention in the robotics community. 
These algorithms are probabilistically complete~\cite{perez}. One of the problems with sampling-based algorithms is that they do not necessarily return a global optimal path. However, there is a variant of RRT that is called RRT*, which finds a path that exponentially approaches the global optimal path in the environment as the number of samples approaches infinity~\cite{sertac}. 

In this paper, it is assumed that a fairly simple 3D representation of the environment in which the vehicle is flying is available a priori. There are obstacles including all sensitive regions in the environment such as the buildings, trees and lakes which we want to avoid colliding with. Obstacles are assumed to be stationary and cuboid. An example of such representation can be found in Fig~\ref{fig:simres}. 

For the given map, using Generalized Voronoi Diagram~\cite{corke} and defining a cost function, the minimum cost landing spot is found. Using RRT* algorithm an obstacle-free path is found to connect the start point to the landing spot. Finally, using a simple search algorithm the path is shortened further (if possible) and the vehicle performs emergency landing by following it. 

\subsection{Finding The Best Landing Spot}
Selecting the location of landing is an important step in emergency situations, simply because it determines the feasibility of the landing. For example, using the distance of the landing spot from the vehicle and a model to compute total power consumption while following a path, one can determine if the vehicle can safely get to its destination. Also, during the path, the vehicle should maintain a certain distance from the obstacles so that in case of complete power outage it would not collide with any of them. In this paper, our goal is to find the best landing spot based on two criteria: (i) finding the safest landing spot with the largest clearance from the obstacles; and (ii) finding the most energy-efficient trajectory towards the landing spot. 

It is assumed that all obstacles are treated the same, therefore the safest way (in terms of collision) to define clearance is to stay at equal distance from them (if possible). One of the best ways to find such points in a map is to use Generalized Voronoi Diagram (GVD). 
Note that because the z-component of the landing spot is always zero (assuming we always land on the ground), the search only takes place in the x-y plane of the given map. However, in order to make this exhaustive search possible, the map is discretized with a step size which is assumed to be 1 meter in this paper (for larger maps or scaled maps, this step size can be scaled to reduce computation time accordingly). 

To find the best landing spot using GVD, a network of obstacle-free paths (edges of GVD) in the x-y plane of the given map is generated. For each point in this network, a cost J as a function of clearance from obstacles and distance from the vehicle is calculated as follows: 
\begin{align}
    \label{eq:path_cost}
    J(r,d) = \frac{a}{r} + bd,
\end{align}
where a and b are two weights to be determined for the clearance from obstacles r and distance from the vehicle d respectively. Finally, by calculating~\eqref{eq:path_cost} for all points in GVD, the point with minimum cost can be selected as the best landing spot. If multiple points are returned, the priority is given to the one with minimum distance from the vehicle.   

\subsection{Path Planning}
In this paper, RRT* algorithm is used to find the path connecting the position of the vehicle to the landing spot in the given map of the environment. In particular, two different scenarios for path planning are evaluated in this paper: (i) When the number of samples are given; and (ii) When the number of samples are unknown. In the first scenario, when the number of samples are given (i.e., 2000 samples), first the graph is generated and then the algorithm attempts to find the shortest path between the start and goal states within that graph (if any exists). Note that in this scenario there is a probability, depending on the number of samples, that the algorithm fails. 

The second scenario is slightly different. Instead of using a fixed number of samples, the algorithm keeps adding vertices to the graph until it finds a path between the start and goal states. As the algorithm adds more vertices, the probability of finding a path between the two points approaches 1 and as the number of vertices approaches infinity, the probability of finding the optimal path approaches 1 as well. An additional step is also added to RRT* which minimizes the length of the path further if possible. Due to the random nature of these algorithms, the final path has unnecessary zig-zag like segments which increases the overall length of the path. To avoid these, a search over the vertices on the final path is performed to find the shortest path among its vertices connecting the start state to the goal state. Either of these two scenarios can be used to plan a path for emergency landing of the quadcopter. 


\section{Simulation Results}\label{sec:sim}
\begin{figure}[t]
	\centering
    \includegraphics[scale=0.3]{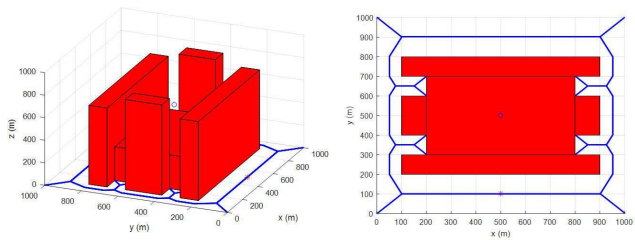}
	\caption{Obstacles are shown in red, GVD is represented by blue lines, initial position of the vehicle is represented by blue circle and the best landing spot is represented by magenta asterisk.}\label{fig:simres} 
\end{figure}

This section is the culmination of all the previous sections and presents simulation results for emergency landing of a quadcopter with one rotor failure. A vehicle with the same specifications introduced in Section~\ref{subsec:tilting} is used in the simulations. 
Suppose that for a quadcopter, motor number 4 is failed (see Fig. 1) and using~\eqref{eq:reduced_atti}, an LQR controller, with $\textbf{Q}=\textrm{diag}(10,1,20,20)$ being the weight matrix for reduced attitude states and $\textbf{R}=\textrm{daig}(5,1)$ being the weight matrix for the control inputs, is designed for the optimal hover solution as found in~\eqref{eq:min_power_hover_sol} (note that LQR is used for simplicity and any other type of controller can be used to control~\eqref{eq:reduced_atti}, also if $\rho=0$, $u_2$ becomes zero too). For position control, damping ratio $\xi=0.65$ and natural frequency $\omega_n=0.8$ are selected for all $x$, $y$ and $z$ coordinates. Representation of the environment is shown in Fig~\ref{fig:simres}, with obstacles being in red. The initial position of the vehicle after failure is at $\pmb{d}_0=(500,500,550)$ m as represented by a small blue circle. By generating GVD for the given map as shown in blue in Fig.~\ref{fig:simres}, and using the cost function defined in~\eqref{eq:path_cost} with weights $a=50000$ and $b=1$, the best landing spot is found to be at $\pmb{d}_{des}=(500,101,0)$ m as represented by a magenta asterisk. 

\begin{figure}[t]
	\centering
    \includegraphics[scale=0.3]{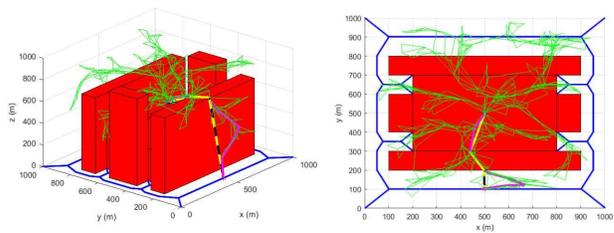}
	\caption{RRT* graph is represented by green, the initial path found by the algorithm is shown in magenta, the final shortest path is shown in yellow and the actual path of the quadcopter is shown in dashed black line.}\label{fig:simres1} 
\end{figure}

Based on the second scenario for path planning and by setting the step size for RRT* algorithm to 50 meters and the radius of the circle to rewire the graph to 150 meters, a path is found between the start and goal states which is shown in magenta in Fig.~\ref{fig:simres} and by searching through the vertices of this path the shortest path can be retrieved as shown in yellow. Finally, by implementing the controller in the nonlinear simulation of the quadcopter flight using~\eqref{eq:rot_motion} and~\eqref{eq:trans_motion}, the vehicle follows the yellow path and lands the vehicle safely. The actual path of the quadcopter following the yellow path is represented by dashed black line in Fig.~\ref{fig:simres1}. 

\begin{figure}[b]
	\centering
    \includegraphics[scale=0.39]{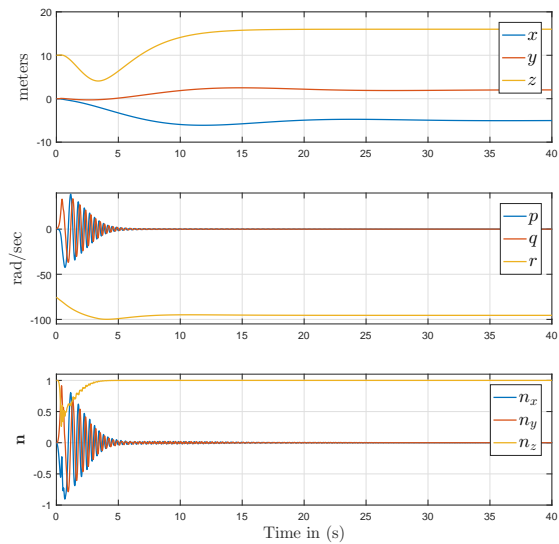}
	\caption{Simulation results for position control. $\pmb{d}_0=(0,0,10)$ m and $\pmb{d}_{des}=(-5,2,16)$ m.}\label{fig:simres2} 
\end{figure}

Another set of nonlinear simulation results, using the same LQR controller for one setpoint change in position of the example quadcopter with the optimal hover solution~\eqref{eq:min_power_hover_sol} is presented in Fig.~\ref{fig:simres2}. Note that the intial yaw rate of the vehicle is chosen to be close to its equilibrium to make sure the linear time-invariant controller is able to stabilize the system. Improving the controller to stabilize the system at arbitrary initial conditions is a topic of future work. 

\section{Conclusions}\label{sec:conc}
This paper presents a framework for emergency landing of quadcopters in case of complete failure of a rotor. Mathematical modeling of a quadcopter considering all significant aerodynamic effects on the propeller's dynamics is presented first. Equilibrium states and fault tolerant control design are presented next followed by introducing a specific configuration for quadcopters which not only results in better stability in yaw motion but also yields the minimum-power hover solution in case of one rotor failure. An algorithm to find the best landing spot using Generalized Voronoi Diagram for a given 3D representation of the environment is introduced for the first time. Finally, path planning for emergency landing using sampling-based planning algorithms (i.e., RRT*) is presented and the results are evaluated by nonlinear simulations. Verifying the results by experiments, investigating the effects of nonzero freestream velocity on spinning vehicles and their power consumption and performing sensitivity analysis can be topics for the future work. 

\bibliography{mybib} 
\bibliographystyle{IEEEtran}
\end{document}